# Unsupervised Deep Representations for Learning Audience Facial Behaviors

Suman Saha
Oxford Brookes University
suman.saha-2014@brookes.ac.uk

Rajitha Navarathna
Disney Research
rajitha.navarathna@disneyresearch.com

Leonhard Helminger
ETH Zürich / Disney Research
leonhard.helminger@disneyresearch.com

Romann M. Weber
Disney Research
romann.weber@disneyresearch.com

## Abstract

*In this paper, we present an unsupervised learning approach for analyzing facial behavior based on a deep generative model combined with a convolutional neural network (CNN). We jointly train a variational auto-encoder (VAE) and a generative adversarial network (GAN) to learn a powerful latent representation from footage of audiences viewing feature-length movies. We show that the learned latent representation successfully encodes meaningful signatures of behaviors related to audience engagement (smiling & laughing) and disengagement (yawning). Our results provide a proof of concept for a more general methodology for annotating hard-to-label multimedia data featuring sparse examples of signals of interest.*

## 1. Introduction

Automatically measuring human behavior is a long-standing challenge in computer vision and machine learning, particularly in the case of detecting emotion or affect in facial expressions [44, 4, 39, 42, 43, 24, 5, 45, 26, 40]. Traditionally, the approach to this problem relies on supervised techniques requiring large amounts of labeled data. However, there is a relative paucity of labeled data in spontaneous, naturalistic settings [44]. As a result, most existing work often centers around exaggeratedly posed behavioral data, usually collected in controlled lighting conditions.

The application of existing techniques is infeasible when applied to large-scale, noisy data featuring multiple faces in poor and inconsistent lighting, such as in a movie-theater setting. Nevertheless, machine learning algorithms capable of capturing audience behavior is of great importance to filmmakers and studio executives during the test-screening process, as it provides for rich, moment-to-moment insight not possible with standard focus-group survey techniques that would be impractical to collect via human annotation for any but the smallest audiences [32].

Several recent attempts have been made to analyze audience reactions within a movie theater [32, 6, 31]. Navarathna *et al.* [32] focused on the body-motion behavior of audience members as measured by optical-flow features, while Deng *et al.* [6] extracted patterns in the dynamics of facial landmarks over the course of a movie using a factorized variational autoencoder (FVAE). Our work is distinguished from this earlier research by not operating on predefined data features (namely optical flow and facial landmarks) but rather on the raw image data of audience members to identify salient facial behaviors such as smiling, closed eyes (or sleeping) and yawning.

With recent advancements in deep feature representations [19], supervised deep-learning frameworks are being increasingly used for facial expression recognition and landmark prediction [37, 14, 22, 23, 21, 13, 12, 43, 29, 30]. However, relatively few attempts have been made to investigate the representational capabilities of unsupervised deep generative models (among the exceptions are [18, 9]). In the present work, we exploit the use of variational autoencoders (VAEs) and generative adversarial networks (GANs) to jointly train a model (VAE + GAN) that incorporates both VAE's reconstruction loss and GAN's binary cross-entropy loss terms to obtain a compact yet expressive latent representation capable of capturing the signatures of various behaviors of interest.

### 1.1. Overview

A schematic providing an overview of our facial analysis framework is shown in Fig. 1. Given video footage of an audience observed during a feature-length film, we predefine a video sub-volume occupied by each audience member (Fig. 1 (**a**)). We then extract key gestures (Fig. 1 (**b**)) for each audience member using template matching (i.e. normalized cross-correlations) as in [31]. Audience key-gesture extraction is useful for identifying potentially "in-teresting" frames, forming a subset many orders of magnitude smaller than the total number of frames. Once the person-specific key gestures have been extracted, we apply a max-margin object detector [17] to extract faces (Fig. 1 **(c)**). We then jointly train a VAE and a GAN model as in [20] (Fig. 1 **(d)**) on these key-gesture images to learn a nonlinear feature embedding from raw pixel space to a latent-space embedding. Finally, we inspect whether the learned representation carries meaningful information about various facial expressions of interest (Fig. 1 **(e)**).

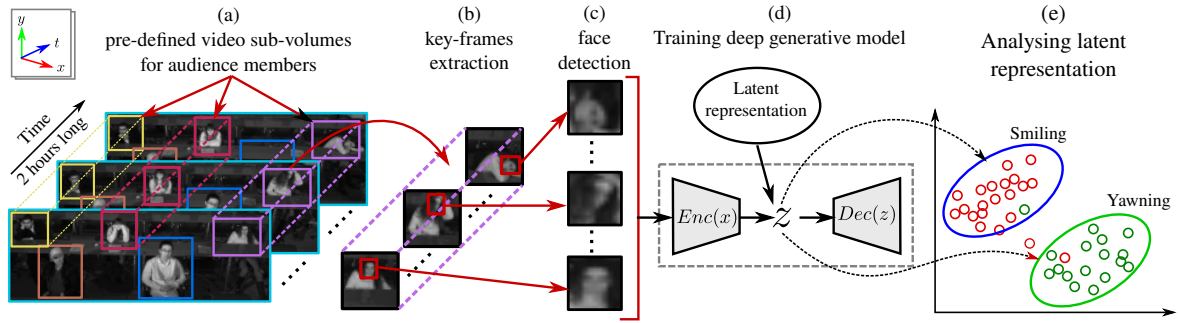

Figure 1. **Overview**. The pipeline consists of (a) predefining a volume for each audience member, (b) extracting the key gestures for each audience member, (c) extracting audience face images using a max-margin object detector, (d) jointly training a VAE and GAN model and (e) analyzing the latent representation.

## 2. Related Work

Conventional methods of estimating the mental state of an audience or viewer sentiment for long-term stimuli, such as movies, stage plays, musical performances, and television shows, are based on self-reports [2, 35]. In movies, where audience responses can be quick and subtle (e.g., a smile at a joke or a jump at a sudden scare), more fine-grained annotation is desirable. However, due to the number of subjects and the long-term nature of the signal, manually annotating audience sentiment is impractically arduous.

Although wearable sensors that gather physiological data (e.g., heart rate, galvanic skin response [3, 8, 33, 27]) or continuous dial ratings [34] could be used, vision-based approaches are ideal, as they can be done unobtrusively and allow viewers to watch the stimuli uninhibited. The first attempt to automate viewer sentiment analysis was proposed by Navarathna et al. [32], where a distribution of short-term correlations of coarse motion features was shown to predict the overall audience rating, although the small size of that dataset limited the generality of the result.

When people watch video clips or listen to music, they often experience an emotional response, which may manifest through various bodily and physiological cues [15, 36]. These cues can often be detected by the trained—and even untrained—eye. Training an *algorithm* to pick up on signals of sentiment, however, poses a considerable challenge. Much of what is known about audience sentiment is volunteered by the audience itself as a self-report. The use of self-reporting for sentiment analysis is not only subjective and labor intensive but also loses much of the fine-grained detail of the sentiment's "dynamics," which can easily be forgotten by a subject by the time the self-report is made.

Joho et al. [11] showed that observed facial behavior is a useful measure of engagement, and Teixerira et al. [38] demonstrated that smiling is a reliable feature of engagement. McDuff et al. [28] further demonstrated the use of smiles to gauge a test audience's reaction to advertisements, while Whitehill et al. [41] used facial expressions to investigate student engagement in a classroom setting. In particular, Whitehill et al. [41] showed that human observers reliably conclude that another person is engaged based on head pose, brows position as well as eye and lip motions.

A large body of work is dedicated to facial expression detection [4, 39, 42, 25, 24, 5, 45]. Most approaches rely on the extraction of facial action units or facial landmarks [7] as input data, with annotations of the depicted expressions provided by human raters operating on the original images. Our approach differs in that it operates on the raw image data rather than predefined features, allowing for potentially richer and more useful features to be automatically learned during training. [1]

## 3. Methodology

The main aims of the work we describe here are twofold: (1) Learn a latent representation of audience images sufficiently rich to faithfully reconstruct them, and (2) isolate "signatures" within the latent representations that correspond to behaviors of interest. In principle, a standard autoencoder could achieve both of these aims. In practice, however, standard autoencoders can become so sensitive to the training data that it can be difficult to determine whether the learned representations correspond to genuinely interesting features in the original data space. It is also often the case that the latent space induced by such models are difficult to explore and sample from.

---

[1]While other recent work employed facial analysis of raw images [1], that work relied on a large, fully annotated dataset, while our method is primarily driven by un- and semi-supervised methods.

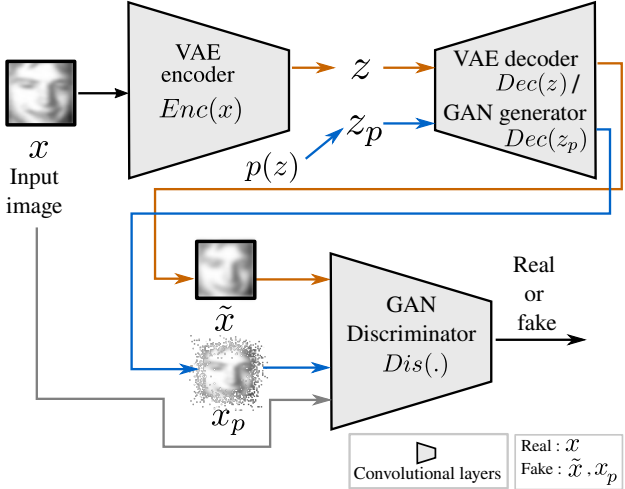

Figure 2. **The VAE+GAN model:** Raw input data, $x$, is encoded into a latent representation, $z$, which is then decoded into a reconstruction, $\tilde{x}$. Separately, random noise, $z_p$, is decoded into a "fake" image, $x_p$. The GAN loss tunes the network so that the noise model generating $z_p$ structures the model so that randomly generated images are indistinguishable from real images.

With these concerns in mind, we deployed the machinery of *generative* models, despite our being primarily interested in latent encodings of existing data. We jointly trained a VAE and a GAN as in [20], to which we refer the reader for details about the architecture, to learn a latent representation of behavior from audience faces. A schematic of the jointly trained VAE and GAN model is depicted in Figure 2.

The VAE and GAN are jointly optimized by the combined objective

$$\mathcal{L} = \mathcal{L}_{\text{prior}} + \mathcal{L}_{\text{like}} + \mathcal{L}_{\text{GAN}}, \quad (1)$$

where, $\mathcal{L}_{\text{prior}}$ is the Kullback-Leibler divergence between the variational approximate posterior distribution $q(\mathbf{z}|\mathbf{x})$ and the prior $p(\mathbf{z})$. That is,

$$\mathcal{L}_{\text{prior}} = D_{\text{KL}}(q(\mathbf{z}|\mathbf{x})||p(\mathbf{z})) \quad (2)$$

The reconstruction error $\mathcal{L}_{\text{like}}$ is the VAE expected log likelihood expressed in terms of the GAN discriminator, namely

$$\mathcal{L}_{\text{like}} = -\mathbb{E}_{q(\mathbf{z}|\mathbf{x})}[\log p(\text{Dis}_l(\mathbf{x})|\mathbf{z})], \quad (3)$$

where $p(\text{Dis}_l(\mathbf{x})|\mathbf{z})$ is formulated as a Gaussian observation model, $p(\text{Dis}_l(\mathbf{x})|\mathbf{z}) = \mathcal{N}(\text{Dis}_l(\mathbf{x})|\text{Dis}_l(\tilde{\mathbf{x}}), \mathbf{I})$, with $\text{Dis}_l(\mathbf{x})$ being the $l^{th}$ hidden layer of the discriminator[2] and $\tilde{\mathbf{x}}$ being the output of the decoder $\text{Dec}(\mathbf{z})$. The hidden layer $\text{Dis}_l(\mathbf{x})$ is itself a representation that corresponds to a learned *similarity metric*; for details, the reader is referred to [20]. The GAN objective is given by

$$\begin{aligned}\mathcal{L}_{\text{GAN}} = -\,[&\log(\text{Dis}(\mathbf{x})) + \log(1 - \text{Dis}(\text{Dec}(\mathbf{z_p})))\\&+ \log(1 - \text{Dis}(\text{Dec}(\text{Enc}(\mathbf{x}))))]\,,\end{aligned} \quad (4)$$

where $\text{Dis}(\cdot)$ indicates the probability that an input is "real" according to the discriminator.[3]

Two main qualities of the VAE+GAN approach inform our analysis. First, by regularizing with a standard normal prior, the VAE objective encourages the model to learn a representation with *independent* components with expected values near zero. Second, in addition to encouraging sharper results from the decoder, the GAN objective effectively *reinforces* the prior constraint by encouraging standard normal noise to decode into plausible images with the characteristics of the dataset. As recent work has argued that additional weight on the prior-matching part of the VAE objective leads to learning *disentangled* representations [10], we suggest that this model formulation achieves a similar aim, albeit through a different route. As we show below, learning disentangled representations is particularly useful when looking for behavioral signals of interest.

## 4. Experiments

We used footage of volunteer audiences watching feature-length movies (roughly 90–120 minutes long). The footage was captured at 15 frames per second with a resolution of $1936 \times 1456$ pixels with infrared-sensitive cameras and was filtered to eliminate flicker from the movie screen [32]. A total of 237 participants were assigned to moderately sized groups (15–25) and were shown one of 10 movies from the animation, comedy, and family genres.

Although not strictly necessary for the present analysis, we winnowed the video data by discovering "key gestures" within a predefined image region for each audience member as in [31]. Intuitively, the method functions as a type of dictionary learning for observed behaviors. We identified over 10,000 such key frames across our audience data. These key frames include expressions such as *looking away*, *smiling*, *yawning*, *closed eyes*, and *neutral* as labeled by human raters.[4] We used max-margin object detector [17] to isolate the faces in these key frames and the Dlib C++ library [16] to train the face detector. Any missed faces were manually cropped for our experiments.

---
[2]There is some freedom of choice here, but see [20] for details.
[3]For more on the GAN objective, see [9].
[4]Note, that these labels were not used during model training.

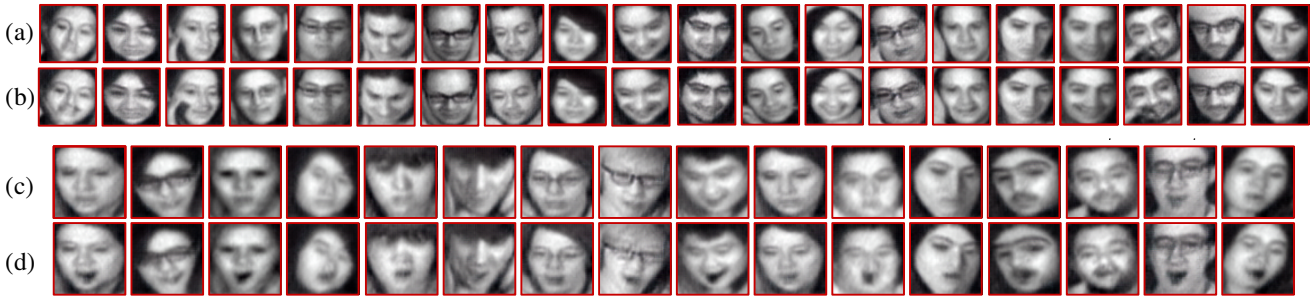

Figure 3. **Attribute vectors and reconstructed images:** Rows **(a)** and **(c)** are reconstructed neutral faces, while rows **(b)** and **(d)** are reconstructed faces after adding *smiling* and *yawning* attribute vectors, respectively, to the neutral faces' latent representations.

### 4.1. Facial Behavior from Latent Representations

There are two main ways for us to examine "interesting" behaviors as captured by latent representations. The first is to observe that all frames for a given subject, indexed by $i$, will have similar encodings, $\mu_i = \frac{1}{T_i}\sum_{t=1}^{T_i} \mathbf{z}_i(t)$; any departure from this expected value, such that $\|\mathbf{z}_i(t) - \mu_i\| > \varepsilon$ for some $\varepsilon$ at time $t$, flags the corresponding frame for further inspection. The second way leverages the labels we have available from human annotators, described below. The interpretation of the results remains the same.

### 4.2. Vector Operations in Latent Space

Of particular interest is the observation that *directions* in latent space are meaningful. That is, a signal corresponding to a given behavior—say, smiling—takes the form of a vector that is shared across encodings [20]. Said another way, the "smile vector" is essentially the same for everyone, and such is the case for the other attributes of interest.

These attribute vectors are isolated by simply averaging over all encodings of images containing the attribute and subtracting off the average of encodings of images *without* the attribute.[5] *Adding* an attribute vector to an encoding of a neutral face introduces that attribute to the decoded image. See Figure 3 for an intuitive example of these attribute vectors in action, demonstrated for the "smile vector" and the "yawn vector."[6]

In identifying these attribute vectors, we are defining a signal-detection problem whose solution leads to the automated annotation of large video datasets. For frames such that $\mathbf{z}_i(t) - \mu_i \approx \mathbf{z}^a$, where $\mathbf{z}^a$ is a given attribute vector, the suggestion is that subject $i$ exhibits the attribute at time $t$.

We can also compute the dot product between an attribute feature vector $\mathbf{z}^a$ and a latent feature vector $\mathbf{z}_i(t)$, with the attribute vector acting here as a *matched filter* to increase the chance of detecting its presence in an encod-ing. Figure 4 shows the result of this operation using the "smile" and "yawn" vectors, showing a clear difference between samples with and without the given attribute.

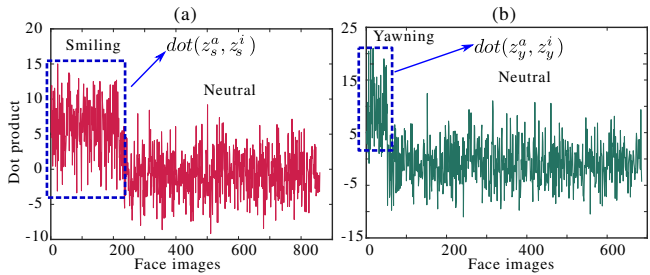

Figure 4. **Dot products in latent space:** Dot products between an attribute feature vector—either (a) *smiling*, $\mathbf{z}_s^a$, or (b) *yawning*, $\mathbf{z}_y^a$—and a latent feature vector $\mathbf{z}^i$.

In the case shown in Figure 4, attribute vectors were isolated by differencing the with-attribute and without-attribute encodings across the dataset, versus subtracting off the *individual* without-attribute means. Although the alternative approach could have led to a better signal-to-noise ratio, we sought to explore the approach in the more likely scenario of having few data labels.[7]

## 5. Concluding Remarks

In this paper we presented a novel application of unsupervised training of generative models to the problem of identifying facial behaviors of interest in a large video dataset. We suggest that this method is especially useful for isolating latent representations of given behaviors that are shared across individuals' encoded data.

As this work was based on an exploratory analysis of our data, future work will leverage our findings to pursue representations with additional favorable properties, particularly with regard to maximizing behaviorally relevant signal against encodings of static features, such as identity.

---

[5]If the learned latent distribution is close to the standard-normal prior, simple averaging over attribute-positive examples is often sufficient, since the larger class of neutral-image encodings should be close to zero mean.

[6]Note in particular the whole-face effect of the introduction of the smiling and yawning attributes, particularly around the eyes.

[7]Yet another alternative would be to simply subtract off the individual encoding means, $\mu_i$, irrespective of label, although this would inevitably dampen some useful signal if the attribute is present in the subject's data.